\documentclass[conference,a4paper]{IEEEtran}
\IEEEoverridecommandlockouts

\usepackage[T1]{fontenc}

\usepackage{subcaption}
\usepackage[numbers]{natbib}
\usepackage{amsmath,amssymb,amsfonts}
\usepackage{graphicx}
\usepackage{xcolor}
\usepackage{authblk}
\usepackage{hyperref}

\def\BibTeX{{\rm B\kern-.05em{\sc i\kern-.025em b}\kern-.08em
    T\kern-.1667em\lower.7ex\hbox{E}\kern-.125emX}}
    
\begin{document}

\title{Utilizing Evolution Strategies to~Train Transformers in~Reinforcement Learning
}

\author[1,2]{Maty\'{a}\v{s}~Lorenc\thanks{Corresponding Author: Matyáš Lorenc; Email: lorenc@kam.mff.cuni.cz}}
\author[2]{Roman~Neruda}

\affil[1]{Faculty of~Mathematics and~Physics, Charles University}
\affil[2]{Institute of~Computer Science, Czech Academy of~Sciences}

\maketitle

\begin{abstract}
We explore the~capability of~evolution strategies to~train an~agent with~a~policy based on~a~transformer architecture in~a~reinforcement learning setting. We performed experiments using \mbox{OpenAI’s} highly parallelizable evolution strategy to~train Decision Transformer in~the~MuJoCo Humanoid locomotion environment and~in~the~environment of~Atari games, testing the~ability of~this black-box optimization technique to~train even such relatively large and~complicated models (compared to~those~previously tested in~the~literature). The~examined evolution strategy proved to~be, in~general, capable of~achieving strong results and~managed to~produce high-performing agents, showcasing evolution's ability to~tackle the~training of~even such~complex models.
\end{abstract}

\begin{IEEEkeywords}
Evolution strategies, Transformers, Policy optimization, Reinforcement learning
\end{IEEEkeywords}

\section{Introduction}

The~problem of~reinforcement learning is considered one of~the~most difficult in~the~field of~machine learning. There are many approaches to~solving it~\cite{RL}. Some are based on computing a gradient to~optimize the~objective, some are derivative-free. One such class of~general derivative-free optimization algorithms are evolutionary algorithms~\cite{EA}. Their subclass of~evolution strategies~\cite{ESIntro} has been proved to~be a viable alternative to~gradient approaches for the~(deep) reinforcement learning~\cite{ESforRL}. Although the~gradient approaches generally have better sample utilization, the~evolution strategies are greatly parallelizable. Moreover, evolution strategies have better exploration of~possible solutions and~the~agents trained using these methods are usually more diverse than those trained by the~gradient-based algorithms. They can even incorporate techniques that~vastly improve the~exploration even more, such~as~searching for~novelty instead~of, or~in~addition to~just seeking better performance. This yields novelty search~\cite{NS1, NS2} or~quality-diversity~\cite{QD} algorithms. An~example of~such~algorithm for reinforcement learning, that~is fairly simple, yet~highly efficient, is \mbox{OpenAI-ES}~\cite{OpenAI-ES}.

On a~different note, transformer architecture~\cite{Transformer} is lately the~go-to solution in~the~field of~neural networks and~supervised learning for an ever-growing range of~problems. And~recently, there have been attempts to~reformulate reinforcement learning as a~sequence modeling problem and~to~leverage the~capabilities of~the~transformers in~such tasks to~obtain a~new approach for solving this class of~problems, yielding models such as Decision Transformer~\cite{DecisionTransformer} or~Trajectory Transformer~\cite{TrajectoryTransformer}. The~Decision Transformer was originally introduced as a~model for offline reinforcement learning using a~supervised learning of~sequence prediction, but the~authors claim it would function well even in~the~classical reinforcement learning tasks.

We decided to~subject the~combination of~the~evolution strategies and~the~Decision Transformers to~experiments, and~test the~ability of~derivative-free algorithms to~train this more complicated and~bigger -- compared to~the~simple feedforward models that had been experimented with in~the~literature so far -- transformer architecture. It might be the~case, that for large and~complicated models it may be hard, or~even almost impossible, to~be trained from scratch using an evolutionary approach. Therefore, we wanted to~experiment with first pretraining the~agent using a~supervised learning of~sequence prediction on data generated by a~smaller, potentially weaker model, which can be trained using any arbitrary reinforcement learning method. We are not searching for~use~cases where~we gain an~edge over~existing approaches by~using Decision Transformers trained via~evolution; rather~we aim to~show that~evolution works well even~on~complex models, and~thus~we need not hesitate to~use the~complex models with, e.g., evolutionary reinforcement learning algorithms~\cite{Combination, EvoRainbow}, the~hybrid state-of-the-art reinforcement learning algorithms that~combine gradients and~evolution, performing better than~the~two individually.

The main contribution of this paper is showing that evolution strategies are capable of~training the~transformer architecture in the reinforcement learning setting -- even~though for~large models they require a~huge number of~CPUs to~run efficiently -- which~allows~us to~harness their~exploration power even~for~transformers. We do~so using \mbox{OpenAI-ES}~\cite{OpenAI-ES}, essentially~the~simplest distributional evolution strategy possible -- if~it works well, all the~more advanced algorithms can only~improve its~performance.

In~the~following section, we present the~background for~our~experiments. In~Section~\ref{Experiments}, we introduce our~experiments and~present their~results, whereas~in~Section~\ref{Discussion}, we discuss these~results. Finally, we conclude the~paper in~the~last section.

\section{Background}

\subsection{Evolution Strategies}

Evolutionary algorithms are a~large and~quite a~successful family of~black-box derivative-free optimization algorithms. They consist of~various metaheuristics inspired by nature. One subclass of~such nature-inspired optimization methods are \textit{evolution strategies}. Introduced as a~tool for~dealing with high-dimensional continuous-valued domains~\cite{ESIntro}, evolution strategies work with a~population of~individuals (real-valued vectors). In~each generation (iteration / population in~the~iteration), they derive a~new set of~individuals by somehow mutating (perturbating) the~original population; the~new set is then evaluated (with respect to~a~given objective function), and~a~new generation is formed based on these new evaluated individuals taking into account their fitness (objective function value).

In~order to~use an~evolution strategy as~a~reinforcement learning algorithm, we use agents represented by a~neural network, or, more specifically, by~a~real-valued vector of~the~network weights, as~the~individuals for~the~algorithm. Then the~only thing needed is to~set the~fitness of~each individual as~the~mean return, the~mean cumulative reward of~the~individual agent from~several episodes. Various evolution strategy algorithms were proposed for this purpose~\cite{ESforRL}.

This approach to~reinforcement learning has its disadvantages. For example, in~order to~compute an~individual's fitness, we have to~run the~entire episodes. Another ailment of~this approach is that its sample utilization is low compared to~gradient-based methods. That is, in~essence, we are capable of~extracting more information from a~timestep or~an episode (a sample) using gradients.

However, evolution strategies also offer many~advantages. Many~of~the~evolution strategy algorithms are easily parallelizable, and~much of~the~research in~the~area has been focused on this aspect, yielding us algorithms with a~linear improvement of~performance when more computing power is used~\cite{OpenAI-ES}. And~because evolution strategies are derivative-free algorithms, we can optimize not just classical smooth neural networks, but our models can even contain some discrete subfunctions or~be otherwise non-differentiable.

In this paper, we will be working with \mbox{\textit{OpenAI-ES}}~\cite{OpenAI-ES}, which is a~representative of~Natural evolution strategies~\cite{NES}. The~population is represented by a~distribution over the~agent's (neural network's) parameters, the~distribution being a~Gaussian, whose mean value is our~current solution to~the~given problem and~from which the~offsprings are sampled each generation. These are then evaluated, and~their evaluation is used to~update the~parameters (in our case only the~mean value) of~the~distribution so that we obtain a~higher expected fitness for the~future samples. This update is performed using an~approximation of~natural gradient (whence the~name Natural evolution strategies). In~our case, this is obtained by renormalizing (rescaling) the~update with respect to~uncertainty. (In general, we would have to~compute an inverse of~a~so-called Fisher information matrix, which would then be used to~multiply the~gradient estimate by. However, as demonstrated, e.g., in~a~paper by~\citet{GradientVariants}, in~our case, when we derive the~parameter updates from the~Gaussian distribution with the~same variance for each parameter, dividing by the~variance (the~rescaling w.r.t. uncertainty) achieves approximately the~same result.) The~algorithm is also designed in~such a~way that it is highly parallelizable with interprocess communication kept at~bare minimum. It shares many~similarities with~DeepGA~\cite{DeepGA}, a~descendant of~genetic algorithms, another branch of~the~family of~evolutionary algorithms. The~main difference is that~DeepGA has explicit population instead~of~the~implicit population coded by~the~distribution of~\mbox{OpenAI-ES}. For~more~details or~for~a~discussion of~the~design choices, we refer our~readers to~the~original paper~\cite{OpenAI-ES}.

\subsection{Transformers}\label{transformers}

\textit{Transformers} have surged as the~state-of-the-art neural architecture for numerous tasks of~supervised learning in~recent years. It all started with their introduction as a~new sequence-to-sequence architecture~\cite{Transformer}, as an~alternative to~recurrent neural networks. Since then they have yielded great results in~natural language processing (NLP) tasks, fueling even the~current surge of~chatbots, and~they even got adapted, e.g., to~image recognition~\cite{VisualTransformer}. As a~rule of~thumb, they seem to~have a~great generalization power; the~greater the~larger the~model employed. However, they also require a~lot of~training data to~achieve these great results.

The~transformers interlace classical feedforward layers with self-attention layers. (In practice, there are also residual connections and~layer normalizations to~stabilize the~training.) And~it is the~self-attention to~which the~transformers owe their success. For each element of~an~input sequence, the~self-attention constructs a~``key'', a~``query'', and~a~``value'' using fully connected layers of~neurons. Then, to~compute an~$i$-th element of~an~output sequence, it takes a~combination of~all the~values, each weighted proportionally to~the~product of~the~query belonging to~the~given ($i$-th) position and~a~key corresponding to~the~value -- hence combining the~information from the~whole input sequence to~produce every single element of~the~output sequence, as shown by the~following equation.

\[ output_i = \sum_{j=1}^n \text{softmax}\left( query_i^T \cdot all\_keys \right)_j \cdot value_j \]

We can, of~course, use a~mask and~for every element hide the~part of~the~sequence that is behind the~element, and~so use just the~information that came before in~the~sequence to~derive the~output element. This is called a~causal masking, and~such a~transformer that uses this masking we then call a~causal transformer.

In the~problem of~reinforcement learning, we want the~agent to~choose actions at~individual timesteps, such that it maximizes its return. This can be, however, viewed as a~sequence modeling problem. And~then, naturally, the~best sequence processing architecture available, the~transformer, comes into focus. Thus, the~\textit{Decision Transformer} was introduced~\cite{DecisionTransformer}.

Its main idea is that we want the~agent's policy to~produce an~action based on not just the~last observation, but~on~the~whole history (or the~part which squeezes into the~context window) of~past observations and~undertaken actions. And~to have some way to~affect the~agent's performance, we add a~conditioning on a~\textit{return-to-go}, which is a~return we want to~obtain from a~given step until the~end of~the~episode.

\begin{figure}[ht]\centering
\includegraphics[width=0.48\textwidth]{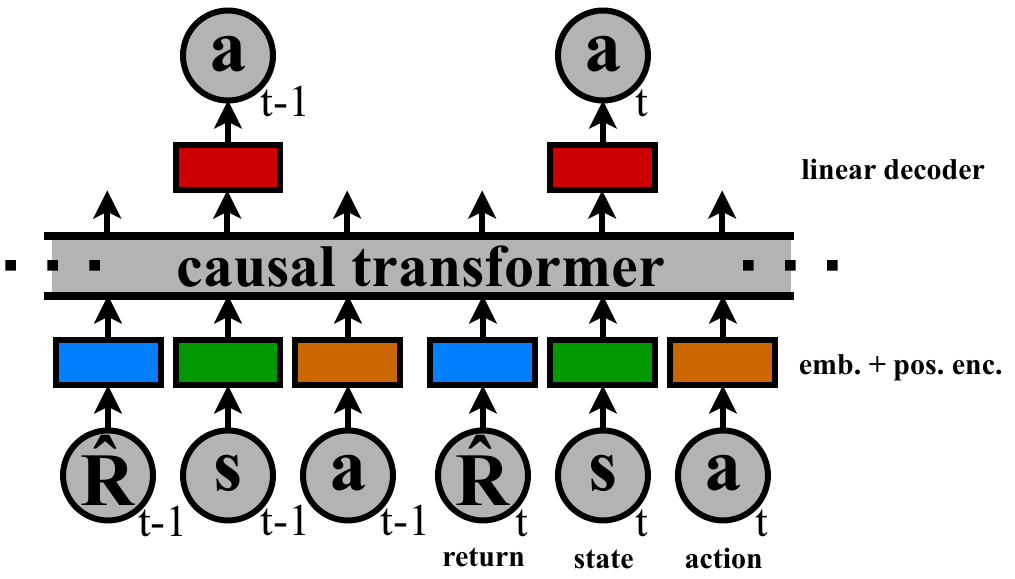}
\caption{Decision Transformer architecture \protect{\cite{DecisionTransformer}}}
\label{DT_architecture}
\end{figure}

So, let us take a~look at~the~proposed architecture itself. The~Decision Transformer consists of~a~causal transformer; embeddings for returns-to-go, observations (states of~the~environment), and~actions; position encoder; and~a~linear decoder to~transform the~output of~the~transformer into~actions, as shown in~Figure~\ref{DT_architecture}.

Every timestep, we feed the~model with a~sequence of~past triplets return-to-go, observation, and~action performed, adding the~current return-to-go and~observation (and a~placeholder for the~yet unperformed action). They get embedded by their respective embeddings, the~positional encoding is added, and~everything is then fed through the~transformer. We then decode the~last output state of~the~transformer to~obtain the~action to~be carried out. An~important difference from the~usual transformer is that every part of~the~timestep triplet, meaning return-to-go, observation, and~action belonging to~one particular timestep, shares one positional encoding, as opposed to~the~classical transformer, where every input sequence element gets its own. Returns-to-go are constructed in~what could be called a~recursive manner. The~user has to~supply the~original one for the~first timestep (which has a~meaning as the~desired performance of~the~model, in~other words target return), while for all the~following timesteps the~return-to-go is constructed by subtracting the~last reward obtained from the~return-to-go belonging to~the~previous timestep.

\subsection{Our \mbox{OpenAI-ES} Implementation}

Mainly due to~different computational resources available to~us -- we needed to~use the~MPI framework instead of~EC2 (AWS) -- as well as the~need to~interchange the~models being trained and~the~environments for the~evaluation more easily and~with a~greater degree of~freedom, we decided to~create our own implementation of~\mbox{OpenAI-ES}.\footnote{Our~code, together with~all the~data gathered during~our~experiments, can be~found on~the~following link: \url{https://github.com/Mafi412/Evolution-Strategies-and-Decision-Transformers}} Yet, in~the~process of~creating this implementation of~ours we noticed some inconsistencies between the~original paper introducing the~algorithm~\cite{OpenAI-ES} and~their provided code. We decided to~adhere more to~the~paper when these dissimilarities occured and~here we state our reasoning for the~two most significant differences.

First and~foremost, in~the~paper~\cite{OpenAI-ES}, the~use of~a~weight decay is mentioned as a~form of~keeping the~effect of~newly added updates still significant enough. After every update, we decay the~weights of~our model, which is the~distribution mean. Therefore, it would not happen that there could be weights grown to~such sizes that a~newly added noise practically does not change the~function of~the~model. This reasoning tells us why we want to~use preciselly the~weight decay. Nevertheless, in~the~original implementation, L2-regularization is used instead. This may be all right for some optimizers, as for, e.g., SGD or~SGD with momentum it is virtually the~same as the~weight decay (just rescaled by the~learning rate). However, as mentioned in~paper \textit{Decoupled Weight Decay Regularization}~\cite{L2vsWeightDecay}, it is not the~same, for example, for ADAM optimizer, which is nonetheless exactly the~one used in~the~original experiments. Our implementation, on the~other hand, remains true to~the~original paper and~uses proper weight decay. Nevertheless, we employ SGD with momentum in~our experiments, as it demonstrated superior efficacy in~our setting during our experiments.

Second, during the~update of~the~distribution parameters, a~gradient estimate should be normalized with respect to~the~uncertainty using a~division by the~standard deviation, so that it becomes a~natural gradient estimate. Despite that, in~the~original OpenAI code, this is not done. There, this is probably hidden in the~value of~a~learning rate (which, as a~result, differs from ours). However, then the~learning rate and~the~noise deviation are unnecessarily coupled hyperparameters.

In addition to~the~implementation of~\mbox{OpenAI-ES}, we made a~few slight modifications even in~the~Decision Transformer code. The~most significant one concerns keeping data from the~past to~be used in~future inputs. In~the~original code, the~authors do not clip sequences already longer than the~context window; those are hence allowed to~grow unconstrained in~the~memory. The~new data are always just appended to~the~list, keeping the~whole history, and~the~whole list is passed to~an~inference function, which only then clips the~sequence to~the~required length (without modifying the~data stored) and~passes it to~the~transformer. This proved to~be no problem in~some environments, but for some reason leads to~a~severe slowdown in~others. Thus, we keep the~stored sequence cropped to~match the~context length in~the~corresponding version of~Decision Transformer.

\section{Experiments}\label{Experiments}

From what we saw in~the~previous section, the~evolution strategies are viable and~well-functioning algorithms for solving the~reinforcement learning problem with~promising parallelization and~exploration abilities~\cite{DRLvsES,DFRL}, whereas the~transformers embody a~new possibility for how to~approach this problem, as well as a~potentially strong architecture of~the~neural network representing the~agent's policy. Therefore, it is just a~logical next step to~try and~combine these two concepts, to~test the~capability of~the~transformers to~be trained by evolution strategies in~the~setting of~reinforcement learning.

As~for~the~evolution strategy, we chose to~use the~\mbox{OpenAI-ES} since~in~its~core it is the~simplest evolution strategy possible. It uses a simple Gaussian distribution with~its parameter covariance matrix being just~a~$\sigma \cdot I$, a~rescaled identity matrix, hence~no~covariance is captured in~this~distribution, nor~any~difference in~significance (or~sensitivity) of~different network weights. It does not even update the~value of~$\sigma$ hyperparameter during training. Therefore, if~this evolution strategy works well, the~others can only improve its~performance.

We provided our own implementation of~the~\mbox{OpenAI-ES} and~began by a~replication experiment of~the~original paper, which also serves as a~correctness check for our code. Here, we tested the~algorithm on a~classical feedforward network in~the~MuJoCo~\cite{MuJoCo} Humanoid environment using OpenAI Gym~\cite{Gym}, whereupon we followed up by proceeding to~test the~performance of~the~evolution strategy with the~Decision Transformer in~the~same environment. We then examined an~idea of~pretraining the~transformer by a~behavior cloning of~some smaller and~easily trained -- yet possibly weaker -- model before training the~transformer by the~evolution strategy. The~last experiments we present are studying the~particulars of~training an~even~larger model in~a~different environment with a~higher-dimensional image input, which is Hero, one of~the~Atari games~\cite{Atari}.

Because training larger models takes more time, and~so~training the~Decision Transformer is particularly time-intensive, we chose the~Humanoid environment as~a~representative of~MuJoCo environments, since it is the~most complex and~challenging among~the~standard ones. As~for~the~Atari, because~there training takes even longer, we selected the~Hero game as~an~example of~a~medium-difficulty Atari game. We use Atari mainly to~demonstrate training behavior on~an~even more complex environment with~a~more~complex visual input, rather~than to~show that~we can solve all its~games.

In~all the~experiments with Decision Transformers, we used the~same hyperparameter values for the~model as used in~the~original paper~\cite{DecisionTransformer} for the~respective environments. For comparison, the~model sizes are 166\,144 parameters for the~feedforward model from the~original \mbox{OpenAI-ES} paper~\cite{OpenAI-ES}, 825\,098 parameters for the~Decision Transformer for MuJoCo Humanoid environment and~2\,486\,272 parameters for Atari games environment.

For~all the~experiments, we conducted ten runs of~the~training. For~every run of~each of~the~experiments, 300 workers were used utilizing 301 CPU cores (with one being a~master handling synchronization, evaluation, and~saving the~agent). Given the~computational complexity of~the~task and~the~fact that~our~goal is to~demonstrate that~the~evolution is~able to~train the~transformer, we~did~not wait for~the~run to~converge. Instead, at~the~beginning, we assigned each~run a~specific number of~iterations it could use to~train the~model, and~once these~iterations were exhausted, we stopped the~training. The~desired returns passed to~all the~Decision Transformer models at~the~beginning of~each episode were 7000 for~the~Humanoid Decision Transformer and~8000 for~the~Atari Decision Transformer. However, here we need to~mention, that~in~the~original Decision Transformer paper~\cite{DecisionTransformer}, for~MuJoCo environments they rescale the~rewards and~returns-to-go dividing them by~1000, probably so~that the~values are somewhat normalized. For~compatibility with~the~paper, we decided to~keep this~rescaling and~hence the~initial return-to-go value passed to~the~model when~7000 is passed as~an~argument to~the~algorithm is~7. All~the~figures reporting fitness progression during~the~evolution training in~Humanoid environment work with~these~rescaled returns as~well, because~the~rescaled values are those~logged during the~training. To~obtain the~fitness value, the~return in~the~form the~original Humanoid environment yields us, we only need to~multiply the~reported value by~1000.

Unless stated otherwise in~the~individual experiment descriptions, all the~other hyperparameter values can be found as~default values in~our codebase.

\begin{figure*}[t!]
	\centering
    \includegraphics[width=0.48\textwidth]{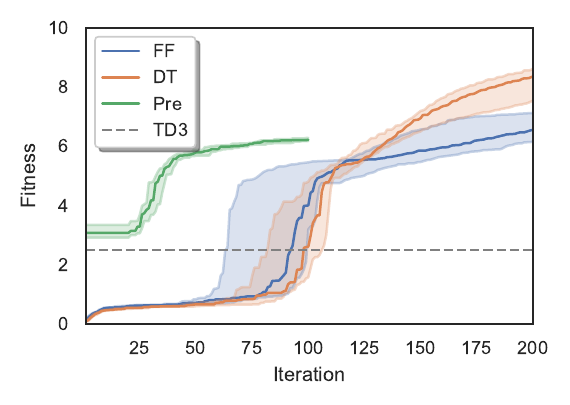}
	\caption{Median and~quartiles of~the~best fitness values so~far for~all the~main experiments done in~the~Humanoid environment for~ten runs each.
	``\textit{FF}'' shows the~training of~a~classical feedforward network using \mbox{OpenAI-ES}, as~described in~Section~\ref{exp_ff_humanoid}.
	``\textit{DT}'' shows the~training of~a~Decision Transformer using \mbox{OpenAI-ES}, as~introduced in~Section~\ref{exp_dt_humanoid}.
	``\textit{Pre}'' shows the~training of~a~pretrained Decision Transformer using \mbox{OpenAI-ES}, as~discussed in~Section~\ref{exp_dt_humanoid_pretrained}. (This~experiment differs from~the~previous two by~the~fact that~its number of~iterations was smaller since~the~model was pretrained, which~is the~reason why~its~line ends in~the~middle of~the~plot.)
	Finally, the~``\textit{TD3}'' horizontal line shows the~final average performance of~the~best Decision Transformer trained by~TD3 in~a~given time, as~mentioned in~the~introduction of~Section~\ref{Experiments}.}
	\label{ES_Humanoid_All}
\end{figure*}

To~have a~baseline to~compare our~experiments with~the~Decision Transformer to, we use the~feedforward model trained using the~\mbox{OpenAI-ES} during our~first experiment described in~Section~\ref{exp_ff_humanoid}. However, to~also~have some~initial comparison to~the~gradient-based methods, we wanted to~try and~train the~Decision Transformer using gradients. The~problem with~this is that~the~Decision Transformer is originally intended as~an~offline reinforcement learning model. There exists a~so-called Online Decision Transformer~\cite{OnlineDecisionTransformer}, but~even it assumes starting with~some~pretraining. Although its~architecture slightly differs from~the~classical Decision Transformer -- it mainly differs in~the~action decoding layer, where a~stochasticity is added -- we tried to~obtain the~gradient baseline by~training it using its~standard training loop, just without~any~initial pretraining on~precollected trajectories. However, the~Online Decision Transformer proved to~be unable to~efficiently improve. Hence, we moved our~attention to~classical online reinforcement algorithms, even though those required a~certain degree of~creativity to~make them work, since~the~Decision Transformer does not take only the~last state as~an~input, but~whole sequences of~recent states, actions and~the~corresponing returns-to-go. Thus, we utilized a~TD3~\cite{TD3}, a~state of~the~art gradient reinforcement learning algorithm, implemented by~Stable Baselines3 library~\cite{StableBaselines3}. We ran the~training for~1\,500\,000 timesteps -- with~this number of~timesteps on~one CPU (environment simulation) and~one GPU (gradient training) it ran for~roughly a~similar wall-clock time as~our~experiments with~\mbox{OpenAI-ES} in~Section~\ref{exp_dt_humanoid}. Still, with~more timesteps, the~model would probably continue to~improve. We conducted five runs of~this gradient training in~Humanoid environment and~selected the~final average performance of~the~best trained model as~a~baseline that is then shown in~Figure~\ref{ES_Humanoid_All}. As~we can see, the~evolution is currently the~only standard online reinforcement learning approach that~can be~applied to~training the~Decision Transformer without~any~adjustment, the~gradient-based ones are either~non-functional or~need to~be~modified first.

\subsection{Feedforward - Humanoid}\label{exp_ff_humanoid}

As stated before, the~first experiment serves as a~replication experiment for the~original \mbox{OpenAI-ES} paper~\cite{OpenAI-ES}, as well as a~sanity check of~our implementation. It consists of~training a~feedforward model using \mbox{OpenAI-ES} in~Humanoid environment. The~hyperparameters are thus mostly directly copied from~the~original paper, with~the~exception of~those affected by~our~implementation changes (e.g.,~a~learning rate) which~were set manually through~experiments. We can see in~Figure~\ref{ES_Humanoid_All} that our implementation is functional and~is comparable in~performance with~the~original implementation. The~progression of~the~training shown in~the~figure also~serves as~a~benchmark against~which the~results of~other experiments can be compared.

\begin{table}[h!]
	\centering
	\caption{Median and quartiles of the average (unrescaled) returns of the trained models with different initial return-to-go argument values.}
	\label{DT_rtg_table}
	\begin{tabular}{r c c c}
	\hline
	Return-to-go & Q1 return & Median return & Q3 return \\
	\hline
	$-1000$ & 7466 & 8447 & 8721  \\
	0 & 7467 & 8440 & 8730  \\
	7000 & 7505 & 8370 & 8672  \\
	1\,000\,000 & 7489 & 8354 & 8667  \\
	\hline
	\end{tabular}
\end{table}

\subsection{Decision Transformer - Humanoid}\label{exp_dt_humanoid}

\begin{figure*}[t!]
	\begin{subfigure}{0.49\textwidth}
        \includegraphics[width=\textwidth]{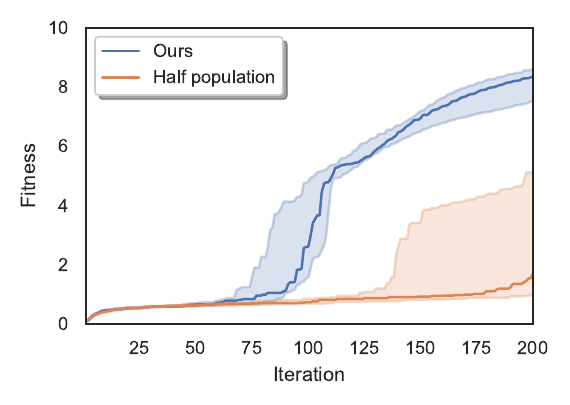}
        \caption{Without pretraining}
        \label{DT_ablation}
    \end{subfigure}
    \hfill
	\begin{subfigure}{0.49\textwidth}
        \includegraphics[width=\textwidth]{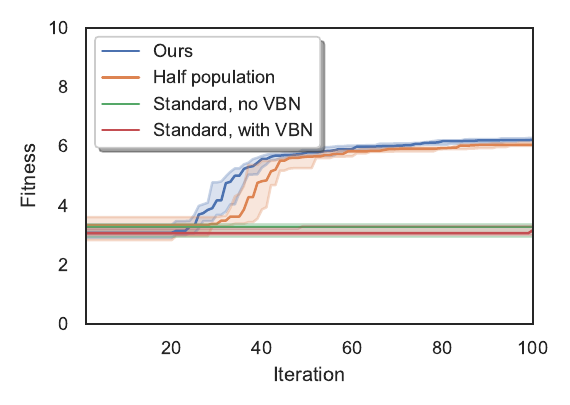}
        \caption{With pretraining}
        \label{DT_pretrained_ablation}
    \end{subfigure}
    
\caption{Results of~the~ablation studies for~training of~Decision Transformers using \mbox{OpenAI-ES} done in~the~Humanoid environment for~ten runs each, aggregated into~medians and~quartiles.
Figure~\ref{DT_ablation} shows the~fitness progression of~the~best yet~found model for~a~training without~first pretraining the~model, as~introduced in~Section~\ref{exp_dt_humanoid}. ``\textit{Ours}'' stands for~our~setting of~hyperparameters; ``\textit{Half population}'' is~the~same except~the~size of~the~population is cut in~half.
Figure~\ref{DT_pretrained_ablation} shows the~\mbox{best-yet} fitness values for~a~training utilizing pretraining, as~introduced in~Section~\ref{exp_dt_humanoid_pretrained}. ``\textit{Ours}'' stands for~our~setting of~hyperparameters; ``\textit{Half population}'' has the~size of~the~population cut in~half; ``\textit{Standard, no VBN}'' has the~same hyperparameters as~when no~pretraining was used, so~the~same as~experiment in~Section~\ref{exp_dt_humanoid}, but~the~virtual batch normalization is not used; ``\textit{Standard, with VBN}'' has exactly the~same hyperparameters as~when~training without~the~pretraining.}
\label{DT_Humanoid_ablations}
\end{figure*}

In this experiment, we inspected the~ability of~the~given evolution strategy to~train even a~more complicated model, which the~Decision Transformers are. We trained a~Decision Transformer model using \mbox{OpenAI-ES}, again in~the~Humanoid environment. Because the~model is almost five times larger, we quadrupled the~size of~the~population the~algorithm works with. The~results are shown in~Figure~\ref{ES_Humanoid_All}. There,~we can see that~the~algorithm is fully capable of~training our~larger model, achieving very good results using a~constant number of~iterations (albeit~at~the~expense of~a~longer wall-clock runtime).

Because a~larger population increases the~time needed for~the~computation with~the~same available resources, a~question arises whether such~increase in~population is truly necessary. Therefore, we also explored whether it is necessary to~use such a~substantial population or~whether half of~it would be sufficient. The~results of~this ablation study can be seen in~Figure~\ref{DT_ablation}. We can see that even though the~algorithm is sometimes still capable of~training the~model, more often than not it does not succeed in~doing so.

Last, we want to~check whether the~trained models respond to~different returns-to-go passed to~the~model at~the~beginning of~each~episode as~the~desired performance, the~target return. Hence, we simulated the~trained model in~the~environment with~various intitial return-to-go values. The~results of~these~experiments can be found in~Table~\ref{DT_rtg_table}. We can see there is no significant difference between the~results of~the~models with~distinct returns-to-go passed.

\subsection{Decision Transformer - Humanoid - Pretrained}\label{exp_dt_humanoid_pretrained}

The~core idea behind this experiment is that if we have a~really large model with many parameters, it might be quite hard for the~algorithm to~gain enough information from just sampling in~a~neighborhood of~a~randomly initialized model. Hence what we might try to~do is first pretrain the~model using behavior cloning towards some smaller, easily trained yet possibly weaker model and~only then utilizing the~evolution strategy to~further improve the~large pretrained model.

Yet, this approach has its costs. First, the~\mbox{OpenAI-ES} uses virtual batch normalization (VBN)~\cite{VBN}. This changes the~inputs to~the~model as the~training progresses. So, either we would have to~use VBN even during the~pretraining, or~we cannot use the~VBN during the~following training with the~evolution strategy. We opted for the~second possibility, as it is more general and~allows us to~showcase a~further training of~any model we might have using the~evolution strategy, even though the~first option would be more favorable for the~algorithm, as the~VBN improves its reliability~\cite{OpenAI-ES} -- it should, however, be reportedly most important when the~model is mostly random at~the~beginning of~the~training.

Second, using the~same values for~learning rate and~noise deviation hyperparameters as~when training from~scratch leads to~a~complete randomization of~the~model at~the~beginning of~the~training and~then the~training proceeds as if from scratch. So, for the~pretraining to~have an~effect, we have to~decrease both values (to 0.01 in~our experiment). This, in~turn, leads to~a~slower progression of~the~training.

Hence, again in~the~Humanoid environment, we pretrained the~Decision Transformer using a~behavior cloning to~a~feedforward model trained using SAC algorithm~\cite{SAC}, before training it further using \mbox{OpenAI-ES}. The~feedforward model and~the~pretrained Decision Transformer had a~fitness value around 4. The~results of~the~subsequent training of~the~pretrained Decision Transformer using the~evolution strategy are shown in~Figure~\ref{ES_Humanoid_All}. Even~though our~approach is capable of~training the~model, the~progression (the~improvement) is visibly slower than~in~the~previous section without~the~pretraining -- except~for~the~initial phase.

Again, we explored the~possibility of~utilizing a~smaller population for this training. As can be seen in~Figure~\ref{DT_pretrained_ablation}, this time, although the~reliability of~the~training is again diminished, most of~the~time the~algorithm works surprisingly well compared to~how it works with the~full population. Still, there~was a~single run in~which the~model did not manage to~improve.

Next, we tried using standard (i.e. the~same as in~the~previous experiment without the~pretraining) values for~the~learning rate and~noise deviation hyperparameters both with and~without the~VBN, as~illustrated in~Figure~\ref{DT_pretrained_ablation}. Without the~VBN the~training showed no progression at~all, and~with the~VBN the~training started to~show similar patterns towards the~end of~the~training in~a~few~cases as~the~one from~Section~\ref{exp_dt_humanoid}, yet~still was maybe a~bit slower.

\begin{figure*}[t!]
	\begin{subfigure}{0.49\textwidth}
        \includegraphics[width=\textwidth]{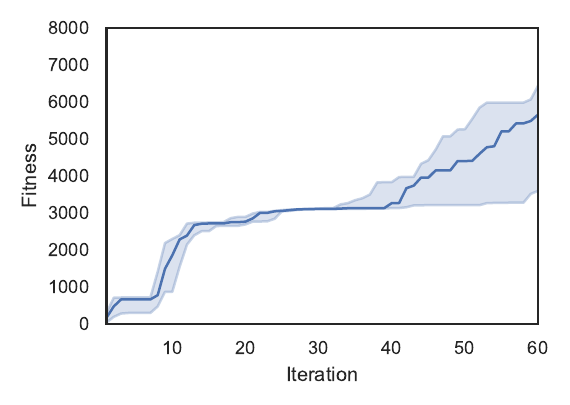}
        \caption{Evaluation results}
        \label{DT_Atari_eval_figure}
    \end{subfigure}
    \hfill
	\begin{subfigure}{0.49\textwidth}
        \includegraphics[width=\textwidth]{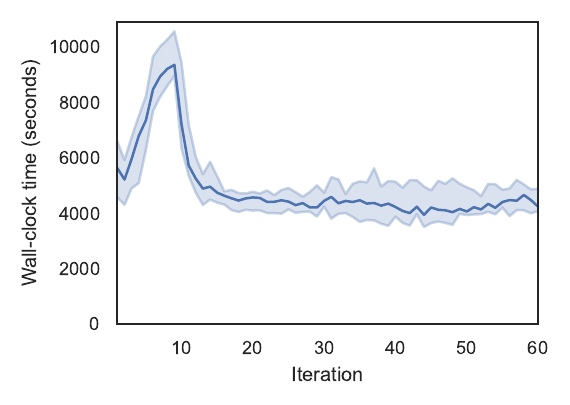}
        \caption{Wall-clock time}
        \label{DT_Atari_time_figure}
    \end{subfigure}
    
\caption{Median and~quartiles of~fitness values and~wall-clock time for~\mbox{OpenAI-ES} used on~a~Decision Transformer model for~Atari game Hero in~ten runs of~the~experiment.
Figure~\ref{DT_Atari_eval_figure} shows us the~best-yet evaluation results after every iteration of~the~algorithm. In~Figure~\ref{DT_Atari_time_figure}, we can see the~wall-clock time the~training needed each~iteration in~seconds.}
\label{DT_Atari_figure}
\end{figure*}

\subsection{Decision Transformer - Atari}\label{exp_dt_atari}

As a~final experiment, we tested the~ability of~\mbox{OpenAI-ES} to~train an~even larger Decision Transformer in~Atari games environment, more specifically, in~the~game \textit{Hero}. Again, since the~model is approximately three times bigger than in~the~previous experiments, we further doubled the~population size for the~algorithm. Even though Figure~\ref{DT_Atari_eval_figure} shows that a~gradual improvement of~the~model really does occur, Figure~\ref{DT_Atari_time_figure} reveals that it is accompanied by quite a~significant increase in~the~time required to~process a~single iteration. This is due to~a~slower model inference, as the~model is substantially larger, as well as longer episodes in~the~environment. (For~the~other experiments, the~wall-clock time is not as~interesting but~can nonetheless be found alongside all the~remaining collected data in~the~GitHub repository.) We did not include a~gradient baseline for~this experiment, as~it primarily serves to~showcase the~much longer training durations required for~such large models.

In Figure~\ref{DT_Atari_time_figure}, we can see an~interesting pattern with a~spike at~the~beginning of~the~training, so let us just note that it is something domain dependent. Here is our attempt to~explain what is most likely going on. At~the~beginning, the~agent performs random actions, which leads it to~repeatedly blow itself up, hence we get relatively short episodes. Then it learns that planting the~bombs randomly is not a~good idea and~stops ending the~episodes prematurely. And~then it gradually learns how to~get further and~further through the~levels of~the~game, but this comes at~the~cost of~it occasionally blowing itself up again, yet this time it at~least serves its progression through the~game.

\section{Discussion}\label{Discussion}

As we can see by~comparing the~experiments in~Figure~\ref{ES_Humanoid_All}, the~larger Decision Transformer model seems to have a~fitness advantage during the~training over the~smaller feedforward model. But what is more important, \mbox{OpenAI-ES} appears to~be capable of~training it without~any~substantial problems, although the~process takes longer. However, the~training is less robust -- during one experiment run, the~model failed to~achieve any~significant improvement. This might have been caused simply by~an~inconvenient seed, when the~perturbations tested during the~training were not suitable. Or~this might be caused by~one~design choice of~the~algorithm, where for~better runtime the~values for~perturbations are pregenerated at~the beginning of~the~training and~so~for~the~larger models we might have to~generate more values at~the~beginning. Or~maybe we might simply need to~further increase the~population size for~greater robustness.

Even though the~parallelization of~the~algorithm significantly improves the~time required by the~algorithm to~train the~model, on such a~massive scale (or even more in~some potential use cases) it still places substantial demands on the~computing hardware, or~rather on the~reliability of~communication between processes.

Regarding our specific usage of~the~algorithm on~Decision Transformers, we need to~settle one thing. As~we introduced in~Section~\ref{transformers}, Decision Transformer operates with~return-to-go tokens, which in~the~original paper~\cite{DecisionTransformer} serve as~a~desired return the~agent should achieve in~a~remainder of~an~episode. This works well when training in~the~original offline reinforcement learning manner, but when using the~evolution strategy to~train the~Decision Transformers in~our~online setting, there is no~pressure to~take these tokens into~account, and~thus the~tokens are then ignored, as can be seen in Table~\ref{DT_rtg_table}. Nonetheless, we believe that only a~minor modification of~the~algorithm may induce the~agent to~pay attention to~these~tokens. Let us propose an~outline of~such~modifications for~a~possible future work. Each evaluation would consist of~several subevaluations; each subevaluation would be assigned a~desired return sampled from a~$\mathcal{N}(\mu,\sigma)$, a~normal distribution centered at~$\mu$ with variance $\sigma$. The~value of~$\mu$ should be computed from~the~returns obtained during the previous iteration in~such~a~manner that it is within~an~interval between~the~best return obtained and~the~mean return obtained during the~last iteration, so~an~improvement is gradually made. The~variance $\sigma$ would be a~hyperparameter, or~it could possibly be a~variation of~the~returns obtained during the last iteration. The~fitness of~an~individual would not be the~mean of~its~returns obtained, as~is the~case in~\mbox{OpenAI-ES}, but~rather a~decreasing function of~the~absolute values of~the~differences between~the~return obtained and~the~desired return in~each subevaluation; ergo, the~larger the~differences, the~smaller the~fitness. This might be, e.g., the~negative of~a~weighted sum of~the~differences.

Now, let us examine the~experiments that utilize the~pretraining described in~Section~\ref{exp_dt_humanoid_pretrained}. As we noted in~the~section~when explaining the~hyperparameter values, the~pretraining does not perform very well. In~order for us to~be able to~use the~pretrained model as a~base for further training -- even if we would be using VBN already during the~pretraining -- we have to~substantially reduce the~ability of~the~algorithm to~further improve the~model, because we have to~change the~learning rate and~noise deviation hyperparameters to~smaller values. Hence, we get slower learning and~less exploration per iteration.

Even when doing so, we can see in~Figure~\ref{ES_Humanoid_All} that the~model first gets ``broken'' -- its performance worsens during first few iterations, the~best-yet found does not improve -- before it starts to~improve again. Nevertheless, the~improvement comes in~terms of~iterations much sooner than when no pretraing is utilized (and it somewhat works even with smaller population). Furthermore, we can observe by inspecting rollouts of~both the~initial and~final models that those two show similar gaits. This similarity gets even more accentuated, when compared to~the~diversity in~final behaviors that are yielded from the~experiment in~Section~\ref{exp_dt_humanoid}, so when the~algorithm is run without the~pretraining. Therefore, we can clearly see that when initialized by the~pretrained model the~training clearly builds upon the~seeded model, despite its initial deterioration.

We hypothesize the~reason for the~above-mentioned deterioration at~the~beginning of~the~training from pretrained model, as well as the~need for the~aforementioned change of~hyperparameters is the~following: Gradient training -- of~which the~behavior cloning is a~representant -- follows a~different overall goal than the~evolution strategy. It strives for the~best possible parameters of~the~model it trains with regard to~its loss function. In~our case of~behavior cloning, this means that we want to~get our model to~behave exactly as the~one we are cloning towards. Nonetheless, this tells us nothing about how the~model behaves when we perform a~slight perturbation of~parameters. Hence, after the~perturbation, the~model is likely to~no~longer function properly -- or~at least as well -- anymore. So we can say that the~model is brittle in~a~sense that even a~slight change of~parameters is likely to~somehow break the~model. This is, however, in~stark contrast to how the~evolution strategy operates and~what it tries to~achieve. The~goal of~the~evolution strategy is not only that its current model performs as well as possible, but that we get the~best possible behavior in~expectation when sampling model parameters from its current distribution. Thus, we believe the~models trained by the~evolution strategy might be considered more robust -- that might help, e.g., in~areas where we train on precise hardware, but the~model needs to~be afterwards deployed on a~cheaper hardware with less precision, where the~weights of~the~model need to~be rounded. Ergo, the~algorithm needs to~first, let us say, ``robustify'' the~model and~only then it can train it towards a~better performance. This~would be~supported by~the~findings of~\citet{Robustness}.

Regardless, in~the~end, the~results shown in~Section~\ref{exp_dt_humanoid} suggest that the~pretraining is not really necessary. But then what explains why the~problem stated at~the~beginning of~Section~\ref{exp_dt_humanoid_pretrained} does not occur in~practice? We think a~possible answer is that~the~algorithm improves not only by shifting model weights towards the~better performing individuals of~the~population, but when there are no really better individuals in the population, it does so mostly by shifting the~weights away from the~worst performing individuals, as the~majority of~the~population consists of~worse individuals. This might indicate why the~algorithm does not really struggle with training larger and~more complex models, where it is even harder to~stumble onto a~working solution just by adding some noise to~the~model parameters.

Finally, let~us just~briefly comment on~Section~\ref{exp_dt_atari}. The~\mbox{OpenAI-ES} is capable of~training even such~larger models, but~the~larger the~model the~more computing power is required for~the~training to~be~concluded in~a~timely manner. Compared to~sizes of~transformer models powering current chatbots, the~Decision Transformer used to~learn the~Atari game is miniscule, and~even~so it required a~lot of~wall-clock time to~undertake just~tens of~iterations of~the~training on~lower~hundereds of~CPUs. Hence,~training really large transformer would~require a~gargantuan number of~CPUs -- but~again, that~is, at~least in~theory, not something unachievable.

\section{Conclusion}

We have investigated the~ability of~\mbox{OpenAI-ES} (and, in~extension, of~similar evolution strategies) to~train larger and~more complex models than the~standard feedforward ones used in~the~literature so~far in~the~reinforcement learning. We have also suggested a~method to~help this training by pretraining the~model using behavior cloning towards some smaller, possibly weaker, yet easier to~train model. However, the~evolution strategy proved to~be capable of~training large models, whereas the~pretraining showed multiple flaws during the~experiments. Still, it helped us shed light onto the~field of~hybridizing gradients and~evolution strategies, as we have discussed based on our experiments. It helped us understand why the~sequential combination of~first training using gradients and~then proceeding to~train using similar distribution-based evolution strategies might not work well. And~it also~allowed~us to~hypothesize that~the~distributional evolution strategies like \mbox{OpenAI-ES} may yield robust models, which~are less~prone, e.g., to~rounding the~weights due to~low-precision hardware they might be deployed~on.

\bibliographystyle{unsrtnat}
\bibliography{bibliography}

\begin{table*}
	\centering
	\caption{Hyperparameter values throughout the experiments (FF - Humanoid Feed Forward; DT - Humanoid Decision Transformer; Pre - Humanoid Decision Transformer Pretrained; Atari - Atari Decision Transformer)}
	\label{Common_hyperparams}
	\begin{tabular}{c c c c c p{0.4\textwidth}}
	\hline
	Hyperarameter & FF & DT & Pre & Atari & Function \\
	\hline
	rtg & N/A & 7000 & 7000 & 8000 & Return-to-go (unscaled) that should be passed to the transformer \\
	size\_of\_population & 5000 & 20000 & 20000 & 40000 & Size of sampled population in each generation \\
	num\_of\_iterations & 200 & 200 & 100 & 60 & Number of iterations / generations the \mbox{OpenAI-ES} will run for \\
	noise\_deviation & 0.02 & 0.02 & 0.01 & 0.02 & Standard deviation value for \mbox{OpenAI-ES}'s distribution \\
	weight\_decay\_factor & 0.995 & 0.995 & 0.995 & 0.995 & Decay factor of weights of the model after each update \\
	batch\_size & 1000 & 1000 & 1000 & 100 & Size of a batch for a batched weighted sum of noises during model update \\
	update\_vbn\_stats\_probability & 0.01 & 0.01 & 0. & 0.01 & Probability of using the data obtained during evaluation to update the Virtual Batch Normalization statistics \\
	optimizer & SGDM & SGDM & SGDM & SGDM & Optimizer used in experiments \\
	learning\_rate & 0.05 & 0.05 & 0.01 & 0.05 & Learning rate (or step size) \\
	\hline
	\end{tabular}
\end{table*}

\end{document}